\pdfoutput=1

\documentclass[11pt]{article}

\usepackage{acl}

\usepackage{times}
\usepackage{latexsym}
\usepackage{graphicx}

\usepackage[T1]{fontenc}

\usepackage[utf8]{inputenc}

\usepackage{microtype}

\title{VISU at WASSA 2023 Shared Task: Detecting Emotions in Reaction to News Stories Leveraging BERT and Stacked Embeddings}

\author{Vivek Kumar \\
University of Cagliari, Italy \\
\texttt{vivek.kumar@unica.it} \\ \\
\textbf{Prayag Tiwari} \\
Halmstad University, Sweden \\
\texttt{prayag.tiwari@hh.se} \\ \And
\textbf{Sushmita Singh}\\
Liverpool John Moores University, UK \\ 
\texttt{sushmitafordata@gmail.com} \\ }

\begin{document}
\maketitle
\begin{abstract}
Our system, VISU, participated in the WASSA 2023 Shared Task (3) of Emotion Classification from essays written in reaction to news articles. Emotion detection from complex dialogues is challenging and often requires context/domain understanding. Therefore in this research, we have focused on developing deep learning (DL) models using the combination of word embedding representations with tailored prepossessing strategies to capture the nuances of emotions expressed. Our experiments used static and contextual embeddings (individual and stacked) with Bidirectional Long short-term memory (BiLSTM) and Transformer based models. We occupied rank tenth in the emotion detection task by scoring a Macro F1-Score of 0.2717, validating the efficacy of our implemented approaches for small and imbalanced datasets with mixed categories of target emotions.
\end{abstract}

\section{Introduction}
Digitalization and ease of access to internet-based intelligent and interactive technologies have led to an unprecedented amount of textual data generation from social media, customer reviews, and online forums. Therefore, the need to accurately understand and extract emotions and sentiments from text has become imperative for two reasons; first, due to their various crucial applications such as sentiment analysis \cite{info14040222}, chatbots, mental health assessment \citep{need4emp}, social media monitoring, market research, brand management, and customer feedback analysis and second to reduce the human efforts, time and resource requirements. The Shared Task on \textit{Empathy Detection, Emotion Classification and Personality Detection in Interactions} of WASSA series 2023\footnote{\url{https://2023.aclweb.org/program/workshops/}}  (previous tasks 2022 \citep{barriere2022wassa} and 2021 \citep{tafreshi2021wassa}) aims to develop models to predict various targets, including emotion, empathy, personality, and interpersonal-index, from textual data \citep{barriere2023wassa}. The shared task consists of five tracks, of which we participated in \textit{Track 3: Emotion Classification} (EMO), which targets emotion classification at the essay level. This work presents two systems to capture the subtle notion of emotions expressed through texts: a) BiLSTM-based \citep{graves2005framewise} DL model using static, contextual, and the combination of static and contextual (stacked) embeddings and b) Bidirectional Encoder Representations from Transformers (BERT) \citep{devlin-etal-2019-bert}. Stacked embeddings \citep{bhandari2022sentiment} are fast-to-train, powerful but underutilized representations; therefore, to reckon their efficacy compared to the transformer model, we have used them in this work. Our proposed systems have performed competitively and got the tenth rank\footnote{The rank is solely based on the submissions done before the deadline of the shared task} in the evaluation phase of the \textit{Track 3: EMO} task. 

The remainder of the paper is structured as follows: Section~\ref{lit_survey} presents the notable research works on emotion detection. Section~\ref{prob_stat} presents the problem statement, dataset description, and the preprocessing strategies applied. In section~\ref{method_material}, we present our different classification systems and the experimental setup. Section~\ref{results_discussion} presents the evaluation results of our proposed systems and comparison with other participating teams of the shared task. Finally, section~\ref{conc} provides the conclusion and discusses the future research directions.

\section{Literature Survey}\label{lit_survey}
The significance of accurate emotion detection and sentiment analysis extends beyond understanding textual data. Recent research has brought the machines one step closer to mimicking humans' innate ability to understand emotional cues from and text and different modalities. Works such as \citep{acheampong2020text,chatterjee-etal-2019-semeval} explore the emotion detection form texts; \citep{zhong2019knowledge,acheampong2021transformer,adoma2020comparative} explores the variants of transformer models useful for emotion detection from texts. Some notable works such as \citep{fi15030110,9746035,bostan-etal-2020-goodnewseveryone,bostan-klinger-2018-analysis,buechel-hahn-2017-readers,sosea2020canceremo} have created the novel datasets from textual and conversational settings to address the scarce data challenges in complex domains for emotion detection. 

\section{Problem Statement, Dataset Description and Data Preprocessing}\label{prob_stat}
In this section, we have mentioned the problem statement tackled, the dataset description, and the data-prepossessing techniques implemented for our experiments.

\subsection{Problem Statement}
In this work, we tackled a multiclass classification problem to predict emotions from essay-level texts. The target labels consist of thirty-one categories of emotions, including individual and mixed sets of emotion categories, as follows: Hope/Sadness, Anger, Sadness, Neutral, Disgust/Sadness, Anger/Disgust, Fear/Sadness, Joy, Hope, Joy/Neutral, Disgust, Neutral/Sadness, Neutral/Surprise, Anger/Neutral, Hope/Neutral, Surprise, Anger/Sadness, Fear, Anger/Joy, Disgust/Fear, Fear/Neutral, Fear/Hope, Joy/Sadness, Anger/Disgust/Sadness, Anger/Surprise, Disgust/Neutral, Anger/Fear, Sadness/Surprise, Disgust/Surprise, Anger/Hope, and Disgust/Hope. 

\subsection{Dataset Description}
\begin{figure*}[!htb]
\includegraphics[width=16.0 cm,keepaspectratio]{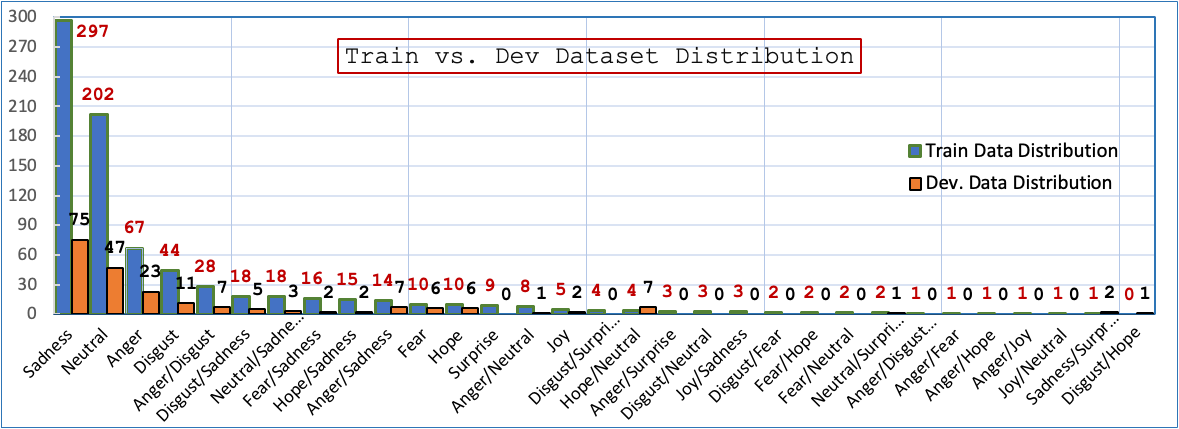}
\caption{Plot showing the skewed distribution of training and development dataset.} \label{train_dev_split}
\end{figure*}

The experimental dataset contains long essays of length between 300 and 800 \citep{omitaomu2022empathic}. The dataset includes news articles and person-level demographic information (empathy, distress, age, race, income, gender, education level, emotion labels, etc.). The dataset was made available as training, development (dev), and test sets where the target labels were shared only for the training and development sets for the evaluation phase. The overall distribution of the dataset is shown in Table~\ref{dataset_split} and the distribution of each emotion class of the train and dev sets is shown in Table~\ref{train_dev_split}.
\begin{table}[!htb]
\centering
\begin{tabular}{llll}
\hline
\multicolumn{4}{l}{\textbf{Dataset Split Distribution}} \\ \hline 
\textit{Train} & \textit{Dev} & \textit{Test} & \textit{Total} \\ \hline 
792 & 208 & 100 & 1000 \\ \hline
\end{tabular}
\caption{Train, dev, and test set distribution.}
\label{dataset_split}
\end{table}

\subsection{Dataset Preprocessing}
As evident from Figure~\ref{train_dev_split}, the dataset is small and imbalanced and several emotion categories have only one data point. Also, the mixed categories of emotions in the target class made the task more challenging. Therefore, to overcome these constraints, we have applied a tailored preprocessing strategy along with standard NLP techniques to prepare the input dataset \citep{DBLP:conf/iui/DessiHKRR20,9286431,uysal2014impact,kumar2019anatomy,info14060349,10.1007/978-3-031-37249-0_10}. The preprocessing steps are as follows. The input texts are converted to lowercase to make the dataset uniform in terms of representation (e.g., Emotion and emotion are represented by a common token, emotion). Punctuation, stopwords, newlines, whitespaces, and extra spaces are removed from the text. We have removed the special characters, symbols, and elements which are not part of the standard English language. We have expanded the contractions such as \textit{didn't} --> \textit{did not}. We performed stemming and lemmatization alternatively for experiments but observed a slight decline in the model's performance. Therefore, we have not considered them for preprocessing the input dataset for the final submission of \textit{Track 3: EMO} shared task.   

\section{Methodology}\label{method_material}
This section describes our different systems (classification models) based on the BiLSTM and transformer model implemented for the emotion classification task. 

\subsection{BiLSTM Based DL Model}
Our first system is a DL model using two BiLSTM layers. More precisely, this model's architecture consists of an embedding layer, followed by two BiLSTM layers, a dense layer, and an output layer at the end for the multi-class classification. The embedding layer is initialized by input\_dim (size of the vocabulary); output\_dim: (word vector length),  embedding matrix, and sequences length. For ease of understanding, we have summed up the parameters and combination of embeddings used for our experiments in Table~\ref{sys_des}. 

\subsection{BERT}
The second system is a transformer-based model \textit{ BERT}, created using \textit{Keras\footnote{https://keras.io/}} and Tensorflow\footnote{https://t fhub.dev/google/collections/bert}. Our \textit{BERT} model comprises two input layers, a \textit{BERT} model layer, and two dense layers of 768 embedding dimensions with the Adam optimizer. The parameters used for fine-tuning the model are listed in Table~\ref{sys_des}. 
\begin{table*}[!htb]
\centering
\begin{tabular}{l|ccccc} \hline
\textbf{Models} & \textit{\begin{tabular}[c]{@{}c@{}}Embedding \\ Dimension\end{tabular}} & \textit{\begin{tabular}[c]{@{}c@{}}Sequence\\ Length\end{tabular}} & \textit{\begin{tabular}[c]{@{}c@{}}Batch \\ Size\end{tabular}} & \textit{Epoch} & \textit{\begin{tabular}[c]{@{}c@{}}Learning \\ Rate\end{tabular}} \\ \hline
\textbf{BiLSTM} + GloVe & 100 & 74 & 32 & 3 & 0.001 \\
\textbf{BiLSTM} + fastText & 300 & 74 & 32 & 4 & 0.001 \\
\textbf{BiLSTM} + (GloVe \& fastText) & 400 & 128 & 32 & 3 & 0.001 \\
\textbf{BiLSTM} + (GloVe \& BERT) & 868 & 128 & 32 & 3 & 0.001 \\
\textbf{BiLSTM} + (fastText \& BERT) & 1068 & 152 & 32 & 7 & 0.001 \\
\textbf{BiLSTM} + (GloVe, fastText \& BERT) & 1168 & 152 & 32 & 5 & 0.001 \\
\textbf{BERT} & 768 & 152 & 32 & 5 & 2e-5 \\ \hline
\end{tabular}
\caption{Experimental settings of proposed systems.}
\label{sys_des}
\end{table*}

\subsection{Features Representation}
We have used pre-trained static and contextual word embeddings for our experiments to generate the feature vectors discussed below.

\textbf{GloVE} (Global Vectors for Word Representation): \textit{GloVE}\footnote{https://nlp.stanford.edu/projects/glove/} is an unsupervised learning algorithm that generates word embeddings as dense vector representations of words in a high-dimensional space. It leverages co-occurrence statistics from a large text corpus to capture semantic relationships between words. \textit{GloVe} embeddings are trained by factorizing a matrix representing the word co-occurrence statistics \citep{pennington2014glove}.

\textbf{fastText}: \textit{fastText\footnote{https://fasttext.cc/}} was developed by Facebook's AI Research (FAIR) team \citep{bojanowski2017enriching,joulin2016bag}. \textit{fastText} extends the traditional word embeddings by representing each word as a bag of character n-grams, where n can range from 1 to a maximum specified length. This approach allows \textit{fastText} to capture morphological information and handle out-of-vocabulary words effectively. 

\textbf{BERT}: \textit{BERT\footnote{https://pypi.org/project/bert-embedding/}} embeddings are a type of word representation that captures contextual information in the text. Unlike traditional word embeddings like \textit{Word2Vec} or \textit{GloVe}, \textit{BERT} embeddings take into account the surrounding words when representing a word. This means that the meaning of a word can vary depending on its context.

\textbf{FLAIR} (FastText and Language-Independent Representations): \textit{FLAIR\footnote{https://pypi.org/project/flair/}} embedding is a state-of-the-art word representation model that captures contextual information and word semantics by combining the strengths of two powerful techniques: \textit{FastText} and contextual string embeddings. By combining these techniques, \textit{FLAIR} embedding provides a robust and language-independent representation of words. It considers both the local context of a word and its global context within a sentence or document \citep{akbik2019flair}.

\section{Results and Analysis}\label{results_discussion}
Table~\ref{my_results} presents the results of our two systems. The BERT base system has significantly outperformed the BiLSTM-based system using the combination of \textit{GloVe}, \textit{fastText} and \textit{BERT} embeddings. Therefore, we have submitted the BERT base result for the shared task evaluation phase. The evaluation of \textit{Track 3: EMO} shared task is based on the macro F1-score and Micro Jaccard-score, Micro F1-score, Micro Precision, Micro Recall, Macro Precision and Macro Recall are supporting metrics. Our BERT base system has achieved a Macro F1-score of 2.717 and stood tenth\footnote{These rankings were provided by the shared task organizing team on 11/05/2023.} among participants. Table~\ref{team_results} presents the official results of all the qualifying teams. 
\begin{table*} [!htb]
\centering 
\begin{tabular}{l|c|c|c|c|c|c|c}
\hline
\textbf{DL Model} & \multicolumn{2}{|c|}{\textbf{Static Embedding}} & \multicolumn{4}{|c|}{\textbf{Stacked Embedding}} & \textbf{Contextual} \\ \hline
\textit{Metrics} & \textit{GloVe} & \textit{fastText} & \textit{\begin{tabular}[c]{@{}c@{}}GloVe +\\ fastText\end{tabular}} & \textit{\begin{tabular}[c]{@{}c@{}}GloVe +\\ BERT\end{tabular}} & \textit{\begin{tabular}[c]{@{}c@{}}fastText\\ + BERT\end{tabular}} & \textit{\begin{tabular}[c]{@{}c@{}}GloVe+\\ fastText\\ +BERT\end{tabular}} & \textit{BERT} \\ \hline
Micro F1-Score & 0.213 & 0.213 & 0.213 & 0.213 & 0.204 & 0.213 & \textbf{0.593} \\
Macro F1-Score & 0.075 & 0.075 & 0.075 & 0.075 & 0.073 & 0.075 & \textbf{0.284} \\
Micro Jaccard & 0.119 & 0.119 & 0.119 & 0.119 & 0.113 & 0.119 & \textbf{0.421} \\
Micro Precision & 0.230 & 0.230 & 0.230 & 0.230 & 0.220 & 0.230 & \textbf{0.640} \\
Macro Precision & 0.046 & 0.046 & 0.046 & 0.046 & 0.045 & 0.046 & \textbf{0.282} \\
Micro Recall & 0.198 & 0.198 & 0.198 & 0.198 & 0.19 & 0.198 & \textbf{0.552} \\
Macro Recall & 0.192 & 0.192 & 0.192 & 0.192 & 0.183 & 0.192 & \textbf{0.318} \\
\hline
\end{tabular}
\caption{The results of our implemented models for static and contextual embeddings.}
\label{my_results}
\end{table*}

\begin{table*}[!htb]
\centering 
\begin{tabular}{c|l|c|c|c|c|c|c|c}
\hline
\textbf{Rank} & \textbf{Team ID} & \textbf{\begin{tabular}[c]{@{}c@{}}Macro \\ F1 Score\end{tabular}} & \textbf{\begin{tabular}[c]{@{}c@{}}Micro \\ Recall\end{tabular}} & \textbf{\begin{tabular}[c]{@{}c@{}}Micro \\ Precision\end{tabular}} & \textbf{\begin{tabular}[c]{@{}c@{}}Micro F1\\ Score\end{tabular}} & \textbf{\begin{tabular}[c]{@{}c@{}}Macro \\ Recall\end{tabular}} & \textbf{\begin{tabular}[c]{@{}c@{}}Macro \\ Precision\end{tabular}} & \textbf{\begin{tabular}[c]{@{}c@{}}Micro \\ Jaccard\end{tabular}} \\ \hline
1 & adityapatkar & 0.7012 & 0.7241 & 0.7778 & 0.750 & 0.6773 & 0.8105 & 0.600 \\
2 & anedilko & 0.6469 & 0.7931 & 0.6259 & 0.6996 & 0.7305 & 0.6305 & 0.538 \\
3 & luxinxyz & 0.644 & 0.6983 & 0.7431 & 0.72 & 0.6314 & 0.7207 & 0.5625 \\
4 & zex & 0.6426 & 0.7069 & 0.7321 & 0.7193 & 0.637 & 0.6992 & 0.5616 \\
5 & lazyboy.blk & 0.6125 & 0.6638 & 0.77 & 0.713 & 0.6005 & 0.7764 & 0.554 \\
6 & gauravk & 0.5649 & 0.7069 & 0.6949 & 0.7009 & 0.5605 & 0.5955 & 0.5395 \\
7 & amsqr & 0.533 & 0.6293 & 0.7228 & 0.6728 & 0.4793 & 0.7521 & 0.5069 \\
8 & surajtc & 0.522 & 0.7586 & 0.5269 & 0.6219 & 0.6679 & 0.4626 & 0.4513 \\
9 & alili\_wyk & 0.5142 & 0.6724 & 0.7358 & 0.7027 & 0.5022 & 0.575 & 0.5417 \\
10 & \textbf{kunwarv4} & 0.2717 & 0.5517 & 0.64 & 0.5926 & 0.3012 & 0.2571 & 0.4211 \\
11 & Cordyceps & 0.202 & 0.4138 & 0.3664 & 0.3887 & 0.2356 & 0.1905 & 0.2412 \\
12 & Sidpan & 0.1497 & 0.4138 & 0.4848 & 0.4465 & 0.2111 & 0.2948 & 0.2874 \\
13 & mimmu3302 & 0.126 & 0.3966 & 0.46 & 0.4259 & 0.2 & 0.092 & 0.2706 \\ \hline
\end{tabular}
\caption{The official results of the evaluation phase of \textit{Track 3: EMO} task. Our system VISU (Team ID kunwarv4} attained the tenth rank.)
\label{team_results}
\end{table*}

\section{Conclusion}\label{conc}
Our system, VISU, participated in the shared task \textit{Track 3: EMO} of emotion classification tasks of the WASSA 2023, and our BERT base system scored tenth rank. Our experiments conclude that although \textit{FLAIR} are powerful word representations built to capture \textit{out-of-vocabulary} words, they are not as effective as contextual embeddings when used for small and imbalanced datasets. Our future research aims to address the data imbalance and scarce data challenges \citep{sdaih23} by incorporating novel augmentation techniques of domain adaptation \citep{9866735} to interpret better the emotions expressed in text. 

\bibliography{biblio}
\bibliographystyle{acl_natbib}




\end{document}